\documentclass[runningheads]{llncs}
\usepackage{graphicx}
\usepackage{hyperref}
\usepackage{listings}
\usepackage{xcolor}

\definecolor{codegreen}{rgb}{0,0.6,0}
\definecolor{codegray}{rgb}{0.5,0.5,0.5}
\definecolor{codepurple}{rgb}{0.58,0,0.82}
\definecolor{backcolour}{rgb}{0.95,0.95,0.92}

\lstdefinestyle{mystyle}{
    backgroundcolor=\color{backcolour},   
    commentstyle=\color{codegreen},
    keywordstyle=\color{magenta},
    numberstyle=\tiny\color{codegray},
    stringstyle=\color{codepurple},
    basicstyle=\ttfamily\footnotesize,
    breakatwhitespace=false,         
    breaklines=true,                 
    captionpos=b,                    
    keepspaces=true,                 
    numbers=left,                    
    numbersep=5pt,                  
    showspaces=false,                
    showstringspaces=false,
    showtabs=false,                  
    tabsize=2
}

\lstdefinestyle{yaml}{
    basicstyle=\ttfamily\small,
    keywordstyle=\color{blue},
    breaklines=true,
    showstringspaces=false,
    identifierstyle=\color{black},
    stringstyle=\color{red},
    commentstyle=\color{gray},
}

\lstdefinelanguage{yaml}{
    morekeywords={true,false,null,y,n},
    sensitive=false,
    morecomment=[l]{\#},
    morestring=[b]",
    morestring=[b]',
    style=yaml
}

\begin{document}
\title{Automated 3D Segmentation of Kidneys and Tumors in MICCAI KiTS 2023 Challenge}
\titlerunning{Auto3DSeg for KiTS 2023}

\author{Andriy Myronenko \and 
Dong Yang \and
Yufan He \and
Daguang Xu 
}

\authorrunning{A. Myronenko et al.}
\institute{NVIDIA \\
\email{amyronenko@nvidia.com}}

\maketitle

\begin{abstract}
Kidney and Kidney Tumor Segmentation Challenge (KiTS) 2023~\cite{kits23} offers a platform for researchers to compare their solutions to segmentation from 3D CT. In this work, we describe our submission to the challenge using automated segmentation of Auto3DSeg\footnote{https://monai.io/apps/auto3dseg} available in MONAI\footnote{https://github.com/Project-MONAI/MONAI}. Our solution achieves the average dice of 0.835 and surface dice of 0.723, which ranks first  and wins the KiTS 2023 challenge\footnote{https://kits-challenge.org/kits23/\#kits23-official-results}.

\end{abstract}


\section{Introduction}

Almost half a million people are diagnosed with kidney cancer annually. Each year, a larger number of kidney tumors are detected, and currently, it is difficult to determine whether a tumor is malignant or benign using radiographic methods. The risk of metastatic progression remains a serious concern, highlighting the need for reliable systems to objectively characterize kidney tumor images and predict treatment outcomes.

For almost five years, the KiTS~\cite{KiTS19Challenge} initiative has maintained and expanded a publicly available collection of hundreds of segmented CT scans featuring kidney tumors. 
%This dataset serves as both a reliable benchmark for 3D semantic segmentation methods and a valuable resource for translational research in kidney tumor radiomics.
This year's KiTS'23~\cite{kits23} competition includes an expanded training set consisting of 489 cases. The goal of the challenge is to develop an automated method to segment kidneys, tumors and cysts.

%###########################
%###########################
\section{Methods}

We implemented our approach with MONAI~\cite{monai} using Auto3DSeg open source project. Auto3DSeg is an automated solution for 3D medical image segmentation, utilizing open source components in MONAI, offering both beginner and advanced researchers the means to effectively develop and deploy high-performing segmentation algorithms.

% Auto3DSeg requires minimal user input, analyzing global data information and automatically generating algorithm folders. 

The minimal user input to run Auto3DSeg for KiTS'23 datasets, is 
\begin{lstlisting}[language=bash]
#!/bin/bash
python -m monai.apps.auto3dseg AutoRunner run \
    --input="./input.yaml"
\end{lstlisting}

where a user provided input config (input.yaml) includes only a few lines:

\begin{lstlisting}[language=yaml]
# This is the YAML file "input.yaml"
modality: CT
datalist: "./dataset.json"
dataroot: "/data/kits23"

class_names:
- { "name": "kidney_and_mass", "index": [1,2,3] }
- { "name": "mass", "index": [2,3] }
- { "name": "tumor", "index": [2] }
sigmoid : true

\end{lstlisting}

When running this command,  Auto3DSeg will analyze the dataset, generate hyperparameter configurations for several supported algorithms, train them, and produce inference and ensembling.  The system will automatically scale to all available GPUs  and also supports multi-node training. 

\begin{figure}[t]
    \centering
    \includegraphics[width=0.8\textwidth]{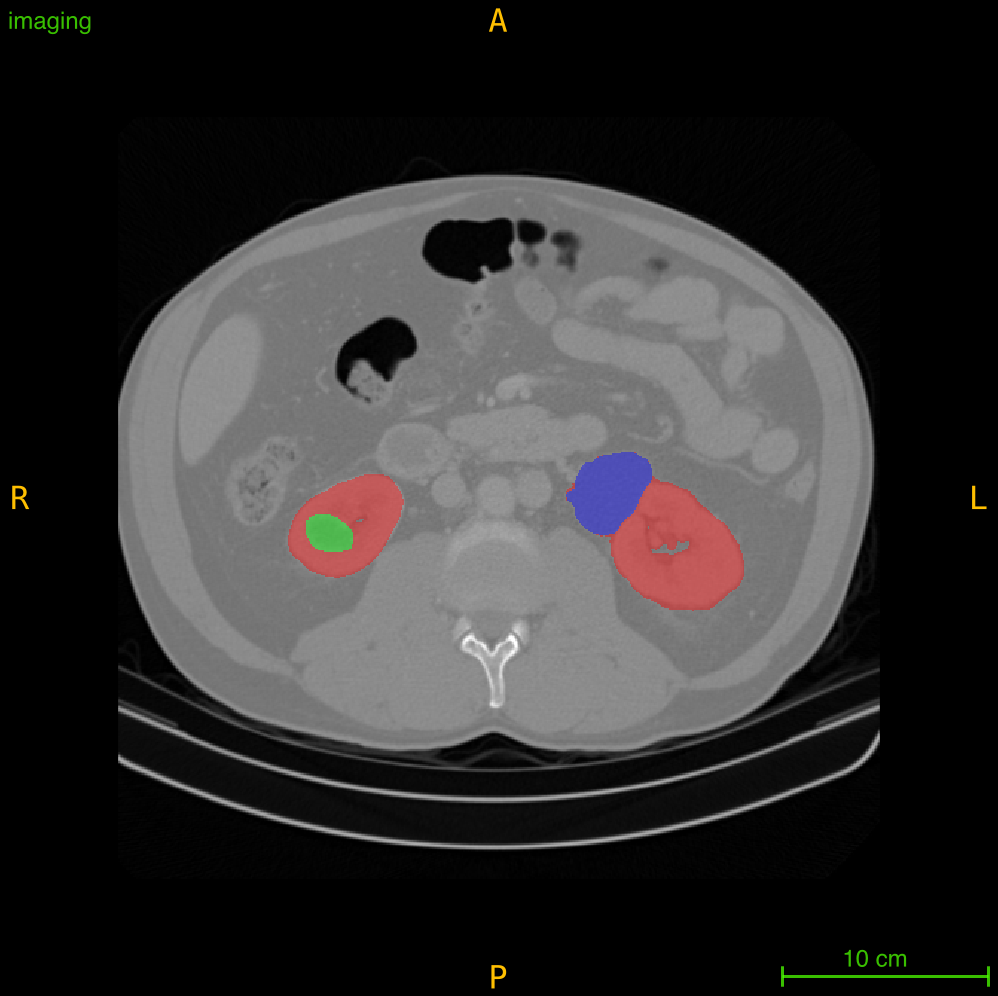}

    \caption{KiTS'23 data example of an axial slice with the provided annotations of kidneys (red), tumors (green) and cysts (blue). The classes of interest that KiTS tasks to segment are: a) all foreground combined b) tumors + cysts (green and blue) c) tumors only (green). }
    \label{fig:example}
\end{figure}

The 3 minimum user options (in input.yaml) are data modality (CT in this case), location of the downloaded KiTS'23 dataset (dataroot), and the list of input filenames with an associated fold number (dataset.json). We generate the 5-fold split assignments randomly. Since KiTS defines its specific label mapping (from integer class labels to 3 subregions, see Fig.~\ref{fig:example}), we have to define it in the config,  and since these subregions are overlapping, we use "sigmoid: true" to indicate multi-label segmentation, where the final activation is sigmoid (instead of the default softmax).

Currently, the default Auto3DSeg setting trains three 3D segmentation algorithms: SegResNet~\cite{myronenko20183d}, DiNTS~\cite{he2021DiNTS} and SwinUNETR~\cite{hatamizadeh2021swin,tang2022self} with their unique training recipes. SegResNet and DiNTS are convolutional neural network (CNN) based architectures, whereas SwinUNETR is based on transformers. Each is trained using 5-fold cross validation.

For model inference, a sliding-window scheme is used to create probability maps, which are re-sampled back to its original spacing. This allows ensembling prediction from different algorithms even if there were trained at different resolutions 

The simplicity of Auto3DSeg is a very minimal user input, which allows even non-expert users to achieve a great baseline performance. The system will take care of most of the heavy lifting to analyze, configure and optimally utilize the available GPU  resources.  And for expert users, there are many configuration options that can be manually provided to override the automatic values, for better performance tuning.

In the final prediction, we ensemble the best model checkpoints only from SegResNet and DiNTS algorithms, since they performed better during cross-validation.  We also applied a few small customizations to the baseline Auto3DSeg workflow. We describe the baseline Auto3DSeg method and the customization below. 

%###########################
\subsection{Training and Validation Data}
Our submission made use of the official KiTS'23 training set alone.

% \subsection{Pre-processing}

% Images are resampled to 0.78x0.78x0.78 mm isotropic resolution,  and re-scaled  from -54..242 to the -1..1 CT interval, followed by a sigmoid. The range was determined automatically by the data analysis step to include the intensity pattern variations within the foreground regions.

% \subsection{Proposed Method}

% We briefly overview the segmentation methods used in Auto3DSeg, which are all freely available in MONAI toolkit.

\subsection{SegResNet}

\begin{figure}[t]
    \centering
    \includegraphics[width=0.8\textwidth]{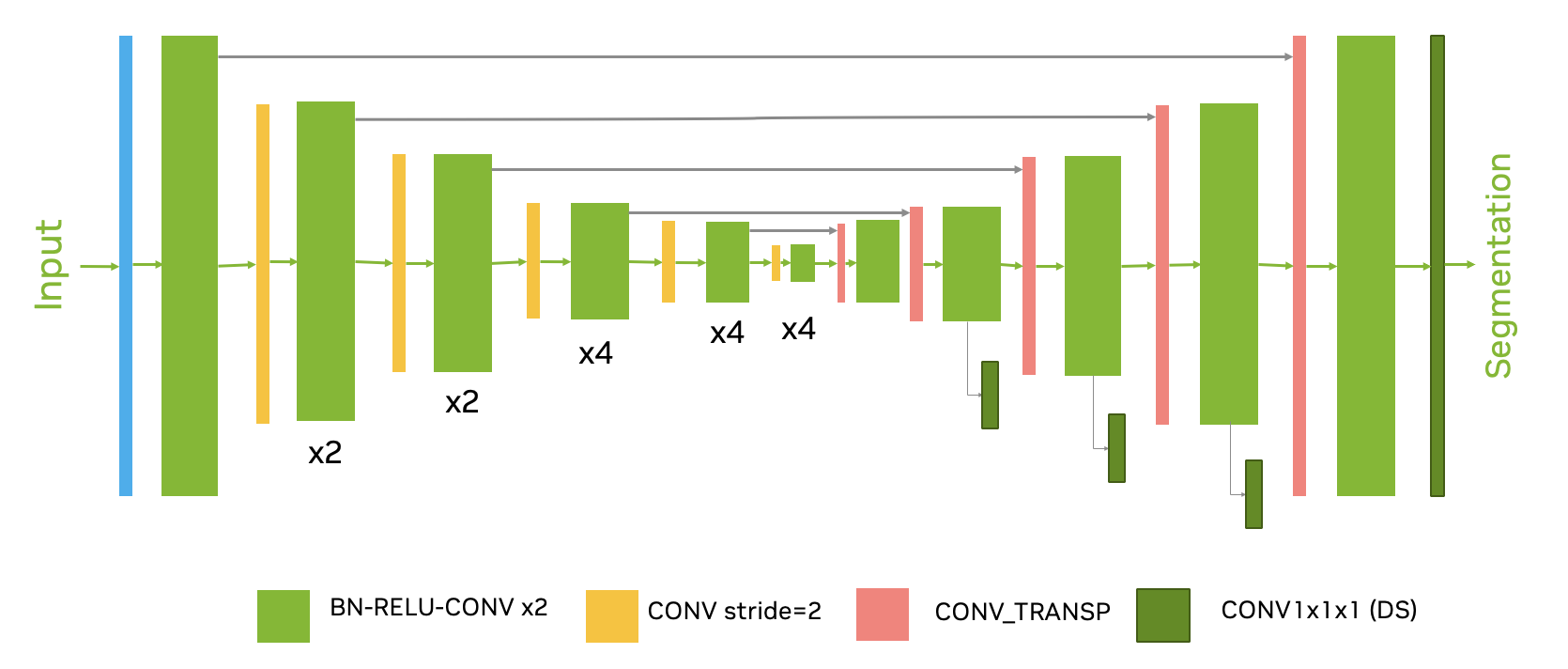}

    \caption{SegResNet network configuration. The network uses repeated ResNet blocks with batch normalization and deep supervision}
    \label{fig:net}
\end{figure}

SegResNet\footnote{https://docs.monai.io/en/stable/networks.html} is an encode-decoder based semantic segmentation network based on~\cite{myronenko20183d}. It is a U-net based convolutional neural network with deep supervision (see Figure~\ref{fig:net}).

The default Auto3DSeg SegResNet configuration was used, which includes 5 levels of  1, 2, 2, 4, 4  blocks.  It follows a common CNN approach to downsize image dimensions by 2 progressively (down to $16 \times$ smaller) and simultaneously increase feature size by.   All convolutions are $3 \times 3 \times 3$ with an initial number of filters equal to 32. The encoder is trained with a $256 \times 256 \times 256$ input region.  The decoder structure is similar to the encoder one, but with a single block per each spatial level. Each decoder level begins with upsizing with transposed convolution: reducing the number of features by a factor of 2  and doubling the spatial dimension, followed by the addition of encoder output of the equivalent spatial level. The number of levels and the region size is automatically configured. We use spatial augmentation including random affine and flip in all axes,random intensity scale, shift, noise and blurring.  We use the dice loss, and sum it over all deep-supervision sublevels:

\begin{equation}
Loss= \sum_{i=0}^{4} \frac{1}{2^{i}} Loss(pred,target^{\downarrow}) 
\end{equation}
where the weight $\frac{1}{2^{i}}$ is smaller for each sublevel (smaller image size) $i$. The target labels are downsized (if necessary) to match the corresponding output size using nearest neighbor interpolation

We use the AdamW optimizer with an initial learning rate of $2e^{-4}$ and decrease it to zero at the end of the final epoch using the Cosine annealing scheduler. We use batch size of 1 per GPU, We use weight decay regularization of $1e^{-5}$. Input images were re-scaled  from $\left [ -54, 242 \right ]$ to the $\left [ -1, 1 \right ]$ CT interval, followed by a sigmoid function. The range was determined automatically by the data analysis step to include the intensity pattern variations within the foreground regions.

\subsection{DiNTS}

DiNTS stands for Differentiable Network Topology Search (DiNTS) scheme, an advanced methodology that fosters more dynamic topologies and an integrated two-level search.
% To minimize the discretization gap during the architecture search process, we propose a guaranteed discretization algorithm along with a discretization-aware topology loss.
% Furthermore, an approach that takes into account memory usage is introduced, which enables the exploration of 3D networks with varying GPU memory requirements.
DiNTS has demonstrated superior performance, achieving top-tier results in the Medical Segmentation Decathlon (MSD) challenge~\cite{antonelli2022medical}.

The DiNTS algorithm utilizes a densely-connected lattice-based network, training with a $96 \times 96 \times 96$ model input for both training and inference.
It leverages automatic mixed precision (AMP) and the SGD optimizer, with an initial learning rate of 0.2 and a loss defined by the Dice plus focal Loss.
For the data processing, we utilize intensity normalization and random  cropping, as well as random rotation, zoom, Gaussian smoothing, intensity scaling and shifting, Gaussian noising, and flipping.

In the quest for enhanced model performance, we fine-tuned the checkpoints that were initially trained using the default training recipe, adjusting them with various patch sizes for a span of 25 epochs.
Our observations suggest that a larger patch size often leads to improved model performance.
Taking into account our computational budget, we have selected a patch size from a range between $192^3$ and $192 \times 192 \times 288$ for each fold of the model (based on validation Dice scores) as the configuration in our final model inference.

% \subsection{SwinUNETR}
% %fill in 
% For the transformer based model, we used the SwinUNETR model~\cite{tang2022self} from MONAI. The input is randomly cropped into 96x96x96 patches with equal sampling rate for each class. 
% %The intensity is cropped at 242.72 and -54.36 (95\% and 5\% intensity interval) and then normalized to 0 and 1.  
% Random flip in all three axis with probability of 0.2, random scale and shift intensity of probability 0.1 are used for data augmentation. For training, we used AdamW optimizer with an initial learning rate of 4e-4 and used the linear warmup cosine scheduler for the learning rate.  

\subsection{Metrics}

The output of the network has 3 channels followed by a sigmoid, to segment each of the 3 KiTS'23 expected classes: a) Kidney + Tumor + Cyst; b) Tumor + Cyst; c) Tumor only.  This creates a multi-label segmentation, where each voxel can belong to more than one label.  We use an average dice metric of these 3 classes to select the best validation checkpoints (without considering surface distances).

\subsection{Auto3DSeg customizations for KiTS'23}

Even though the default Auto3DSeg configuration achieved a good baseline cross-validation performance automatically, we did a few customization including cropping to kidneys region and post-processing. 

Training on the full size 3D CT images can be time consuming, so we pre-cropped the images around the kidneys region. A simple rectangular box was used, based on the ground truth labels.  Since we used only 1 cropping per image, the cropped region included not only kidneys but everything in between including the spine.  Training on such cropped images has 2 advantages: firstly, it allowed for faster training, since smaller images can be cached in RAM, and secondly, it simplified the task for the network.  The disadvantage of such an approach is that it requires finding the bounding box of the kidneys region first.

We trained a separate segmentation network to find the foreground and calculate the kidneys bounding box coordinates.  For all the tasks we used the same exact network architecture, trained all at $0.78 \times 0.78 \times 0.78 mm^3$ CT resolution.  Arguably, bounding box detection could have been done faster, using a simpler detection network and at a lower CT resolution, but here we saved on coding time, by reusing the framework. We trained the first round of models fast (using a smaller number of epochs), to be used as a bounding box detector. And after that trained longer, using only the cropped (around kidneys) regions.  This approach is somewhat similar in spirit to the KiTS 2021 champion solution of coarse-to-fine training~\cite{Zhongchen21} (based on nnU-net~\cite{nnunet21}), but here we do not re-use or concatenate masks detected at a coarse level, we simply use it to detect the bounding box for faster training. 

We also added binary post-processing on the final segmentation masks. Firstly, we remove small connected components (smaller then 100 voxels total) based on the foreground (merged labels). Secondly, we  correct for "outline" of some tumor region.  During a training stage, we noticed that on a small set of images, network predictions of the tumor label have a small rim (1-2 voxels) of cyst label.  This happened mostly because the network was trained as a multi-label task, where each voxel can be assigned to several classes. Since, it's not possible for a tumor to have a "cystic" outline (even if it looks like a cyst image pattern), by definition, we decided to correct such cases with a simple binary post-processing. In our cross-validation tests, this final post-processing did not actually affect the accuracy metrics, but we still decided to include it.

Finally. we increased the size of the network input patch during training to $256 \times 
 256 \times 256$ for SegResNet and to $192 \times 192 \times 288$ for DiNTS, which allowed for faster training  and also slightly increased the cross-validation performance. 

\subsection{Optimization}
We train the method on an 8-GPU 48GB NVIDIA A40 machine, with a batch size of 1 per GPU, which is equivalent to batch size of 8 single GPU training.  Auto3DSeg caches on-the-fly all the resampled data in RAM during the first training epoch, when sufficient amount of RAM is available (otherwise a fraction of the data is cached).  This way only the first epoch suffers a slow-down due to disk i/o and resampling, and the rest of the training process is fast.

\section{Results}

Based on our random 5-fold split, the average dice scores per fold are shown in Table~\ref{tab:result}.  
For the final submission we used an ensemble of 15 models: 10 models of SegResNet (5 folds trained twice), and 5 models of DiNTS.  SegResNet A and B training runs in Table~\ref{tab:result} had the same  configurations. 
\begin{table}[h!]
    \centering
    \begin{tabular}{| c | c | c | c | c | c | c |}
        \hline
        & {\textbf{Fold 1}} & {\textbf{Fold 2}} & {\textbf{Fold 3}} & {\textbf{Fold 4}} & {\textbf{Fold 5}} & {\textbf{Average}} \\
        \hline
        SegResNet A & 0.8997 & 0.8739 & 0.8923 & 0.8911 & 0.8892 & 0.88924\\
        SegResNet B & 0.8995 & 0.8773 & 0.8913 & 0.889 & 0.8865 & 0.88872\\
        DiNTS & 0.8810 & 0.8647	& 0.8806	& 0.8752 & 0.8822 & 0.8767\\

        \hline
    \end{tabular}
    \caption{Avgerage Dice results of the 15 trained models based on our 5-fold data split.}
    \label{tab:result}
\end{table}

On the final hidden challenge dataset, our submission achieved an average Dice score of 0.835, which  ranked first among other submissions.

%###########################
%###########################
\section{Conclusion}
We described our winning solution to KiTS 2023 challenge using Auto3DSeg from MONAI.
Our final submission is en ensemble of 15 CNN models, 10 of SegResNet and 5 of DiNTS. 
We hope that open source tools in MONAI will help more researchers to achieve good baseline 3D segmentation results on their particular task.  Our solution achieves the average dice of 0.835 and surface dice of 0.723, which ranks first on the KiTS 2023 leaderboard\footnote{https://kits-challenge.org/kits23/\#kits23-official-results}.

\bibliographystyle{splncs04}
\bibliography{paper}

\end{document}